\documentclass[11pt]{article}

\usepackage[preprint]{acl}

\usepackage{times}
\usepackage{latexsym}
\usepackage[T1]{fontenc}
\usepackage[utf8]{inputenc}
\usepackage{microtype}
\usepackage{inconsolata}
\usepackage{graphicx}
\usepackage{tikz}
\usetikzlibrary{arrows.meta,positioning,fit,calc,shapes.geometric}
\usepackage{algorithm}
\usepackage{algpseudocode}
\makeatletter
\renewcommand{\theALG@line}{\thealgorithm.\arabic{ALG@line}}
\@ifundefined{theHALG@line}
  {\newcommand{\theHALG@line}{\thealgorithm.\arabic{ALG@line}}}
  {\renewcommand{\theHALG@line}{\thealgorithm.\arabic{ALG@line}}}
\makeatother
\usepackage{wrapfig}
\usepackage{booktabs}
\usepackage{multirow}
\usepackage[table]{xcolor}
\definecolor{imprcolor}{HTML}{EFEBF7}

\usepackage{amsmath}
\usepackage{amssymb}
\usepackage{mathtools}
\usepackage{amsthm}

\usepackage[capitalize,noabbrev]{cleveref}
\usepackage[table, svgnames]{xcolor} 
\usepackage{xspace}
\usepackage{enumitem}
\usepackage{amsmath}
\usepackage{amssymb}
\usepackage{float}

\newcommand{\name}{\texttt{AMREC}\xspace}
\newcommand{\define}[1]{\vspace{0mm}\noindent{{\textbf{#1.}}}}

\usepackage[most]{tcolorbox}
\newtcolorbox{promptbox}[1]{
  colback=gray!10,
  colframe=gray!55,
  coltitle=white,
  colbacktitle=gray!55,
  title=\textbf{#1},
  fonttitle=\bfseries,
  arc=2mm,
  boxrule=0.8pt,
  left=2mm,
  right=2mm,
  top=2mm,
  bottom=2mm
}

\title{Agentic Molecular Recovery via Molecule-Aware Exploration}

\author{Suwan Yoon,\quad Changhee Lee\thanks{Corresponding author.} \\
  Department of Artificial Intelligence, Korea University / Seoul, Korea \\
  \texttt{suwanyoon, changheelee@korea.ac.kr}}

\begin{document}
\maketitle
\begin{abstract}

Text-guided molecular generation with LLMs often yields invalid SMILES. We argue that invalid drafts should be addressed through a shift from validity-oriented repair to identity-preserving molecular recovery: the objective is not only to restore chemical validity, but also to preserve target-relevant structural cues and recover the molecular identity implied by the description. This perspective reveals the limitations of existing correction strategies. Post-hoc repair can recover validity while distorting key structures, LLM-only correction can introduce unintended global drift, and generic agentic correction remains constrained by greedy single-candidate trajectories even when equipped with executable RDKit edit tools. To address these limitations, we propose \name, which couples molecule-aware mismatch tracking with expanded candidate exploration and trajectory-level selection. On invalid ChEBI-20 drafts from three backbone models, \name achieves the strongest overall recovery profile across structural, exact-match, and string-level metrics.

\end{abstract}
\section{Introduction}

Language models are increasingly used for text-guided molecular \textit{de novo} generation, where a model generates molecular structures from natural-language descriptions of desired properties, functions, and chemical structures \citep{edwards2021text2mol,edwards2022translation,li2024empowering}. 
However, standard SMILES-based generation often produces invalid strings that cannot be directly parsed into molecular graphs. 
While invalid outputs are often treated as failures to discard, recent work suggests that they may still encode meaningful chemical information, and that strictly validity-constrained representations can themselves introduce structural bias \citep{skinnider2024invalid}.
This shifts the central question from simply preventing invalidity to restoring chemically meaningful molecules from imperfect drafts.
\begin{figure}[t]
    \centering
    \includegraphics[width=\columnwidth]{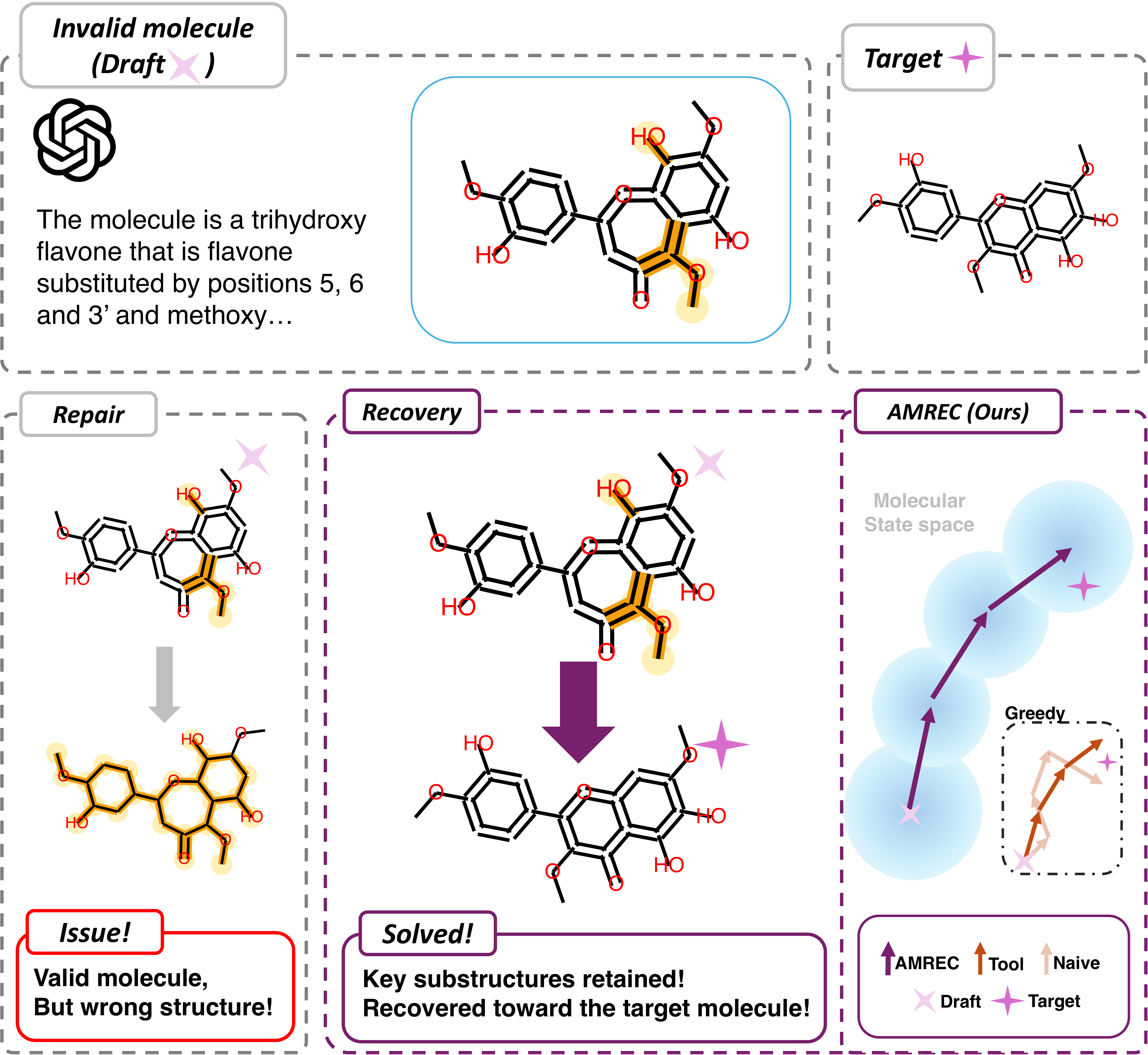} \vspace{-6mm}
    \caption{Conceptual overview of molecular restoration. Repair focus purely on syntax-level validity correction, whereas recovery achieves semantic realignment by preserving target-relevant structural cues and reconstructing the intended molecular identity from the description.}
    \label{fig:figure1} \vspace{-6mm}
\end{figure}

We formalize this problem as \textit{molecular restoration}, distinguishing between two levels of string correction: \textit{Repair} denotes syntax-level validity restoration---simply converting an invalid SMILES into a valid SMILES. \textit{Recovery}, in contrast, denotes semantic and context-aware restoration---restoring chemical validity while preserving target-relevant structural cues in the draft and recovering the molecular identity implied by the target description. 
Existing post-hoc approaches based on rules or validity-preserving representations such as SELFIES can often repair invalid SMILES \citep{krenn2020self,tao2025make}, but validity alone does not guarantee preservation of important structural characteristics, such as scaffolds, ring systems, functional groups, or charge patterns. 
Conversely, LLM-based correction methods can condition on both the target description and validity feedback, yet they typically regenerate entire SMILES strings at each step, making it difficult to separate intended local edits from unintended global structural changes.

This limitation motivates an \textit{agentic} approach to molecular recovery, treating the candidate molecule as the state, factual observations, and structural modifications as actions. Prior tool-augmented molecular agents have addressed \textit{action fidelity} in agentic search by replacing unconstrained full-SMILES regeneration with RDKit-based executable edit tools, so that LLM-selected actions are grounded as deterministic molecular-graph operations \citep{landrum2013rdkit, zhou2026toolmol}. However, tool grounding alone is insufficient for robust molecular recovery. Even with executable editing tools, generic tool-augmented agents remain constrained by weak molecule-aware alignment and greedy single-candidate search, a paradigm we term \textit{agentic greedy search} \citep{yao2022react,xu2023rewoo,erdogan2025plan}. Rather than explicitly tracking molecule-text mismatches as verifiable requirements, these agents greedily refine a single candidate molecule step by step. As a result, even seemingly reasonable local edits may optimize toward the wrong structural objective, while early erroneous decisions can irreversibly steer the recovery into a dead-end trajectory.

\define{Contribution} To address these limitations, we propose \textbf{\name}, an agentic framework for molecular recovery that combines grounded molecular editing with molecule-aware reasoning and expanded candidate exploration. 
\name first derives explicit and verifiable structural requirements from the target description, then employs \texttt{Checker}, \texttt{Critic}, and \texttt{Planner} agents to track semantic mismatches between the current molecule and the intended target. 
Instead of committing to a single greedy trajectory, \name introduces a \texttt{Candidate Explorer} and selection mechanism that builds, retains, and revisits multiple recovery candidates throughout the search process. 
As a result, \name better understands molecular states and explores recovery candidates more effectively, achieving \textbf{the strongest overall recovery performance} on invalid drafts across three backbone models.

\section{Related Work}


\define{LLM-based molecular generation}
Recent studies have explored LLM-based molecular generation through both task-specific training and few-shot prompting.
While trained molecule-language models translate between molecular strings and natural-language descriptions, prompting-based approaches use demonstrations and textual instructions to guide molecular design without task-specific fine-tuning~\citep{edwards2021text2mol, edwards2022translation, pei2023biot5, guo2023can, li2024empowering, li2025large}.
This few-shot paradigm has recently become a major practical direction for applying LLMs to chemistry.
In contrast to these works, which mainly focus on generating molecules from text, our work studies the post-generation recovery of chemically meaningful molecules from invalid molecular drafts.

\define{Validity-oriented molecular repair}
Although SMILES is widely used for molecular generation, its strict grammar often yields invalid outputs \citep{weininger1988smiles}.
At the same time, strict validity enforcement during generation can constrain exploration \citep{skinnider2024invalid}.
This motivates post-generation repair methods such as SMISELF~\citep{tao2025make,schoenmaker2023uncorrupt}.
However, these methods mainly restore validity, rather than semantically recovering the molecular identity implied by the target description.

\define{Chemical agents}
LLM agents combine reasoning, planning, and tool use for iterative decision-making \citep{yao2022react,xu2023rewoo,erdogan2025plan}.
In chemistry, this paradigm has been applied to validation, property evaluation, molecular optimization, and drug discovery with external or executable chemistry tools \citep{m2024augmenting,ansari2024dziner,le2024agentdrug,landrum2013rdkit,zhou2026toolmol}.
However, these works do not address molecule recovery, which requires restoring validity while preserving target-relevant structural cues from corrupted drafts.

\section{Towards Agentic Molecular Recovery}

\begin{table*}[t]
\centering
\scriptsize
\setlength{\tabcolsep}{7pt}
\renewcommand{\arraystretch}{0.9}
\caption{Post-hoc analysis of invalid molecular drafts generated with GPT-5.4-mini at ChEBI-20 dataset. LLM-Corrector is run for five iterations, achieving 40.7\% raw validity; for fair comparison, SMISELF is then applied only to its remaining invalid outputs.}
\label{tab:posthoc_invalid_analysis}
\begin{tabular}{l|c|c|c|c|c|c|c}
\toprule
\rowcolor{gray!18}
Subset / output & MACCS FTS$\uparrow$ & RDK FTS$\uparrow$ & Morgan FTS$\uparrow$ & Exact$\uparrow$ & BLEU$\uparrow$ & ROUGE-L$\uparrow$ & Levenshtein$\downarrow$ \\
\midrule
\rowcolor{gray!10}
\multicolumn{8}{c}{\textit{Initially invalid subset}} \\
Invalid initial draft & -- & -- & -- & -- & 0.6407 & 0.7247 & 46.4433 \\
SMISELF & 0.7154 & 0.3708 & 0.2628 & 0.0000 & 0.4467 & 0.6295 & 50.8814 \\
LLM-Corrector & 0.7505 & 0.4407 & 0.3239 & 0.0567 & 0.4979 & 0.6547 & 49.9948 \\
\midrule
\rowcolor{gray!10}
\multicolumn{8}{c}{\textit{Initially valid subset}} \\
Valid initial draft & \textbf{0.9498} & \textbf{0.8808} & \textbf{0.8336} & \textbf{0.5079} & \textbf{0.8833} & \textbf{0.9222} & \textbf{10.9272} \\
\bottomrule
\end{tabular}%
\end{table*}

\begin{figure}[t]
    \centering
    \includegraphics[width=\columnwidth]{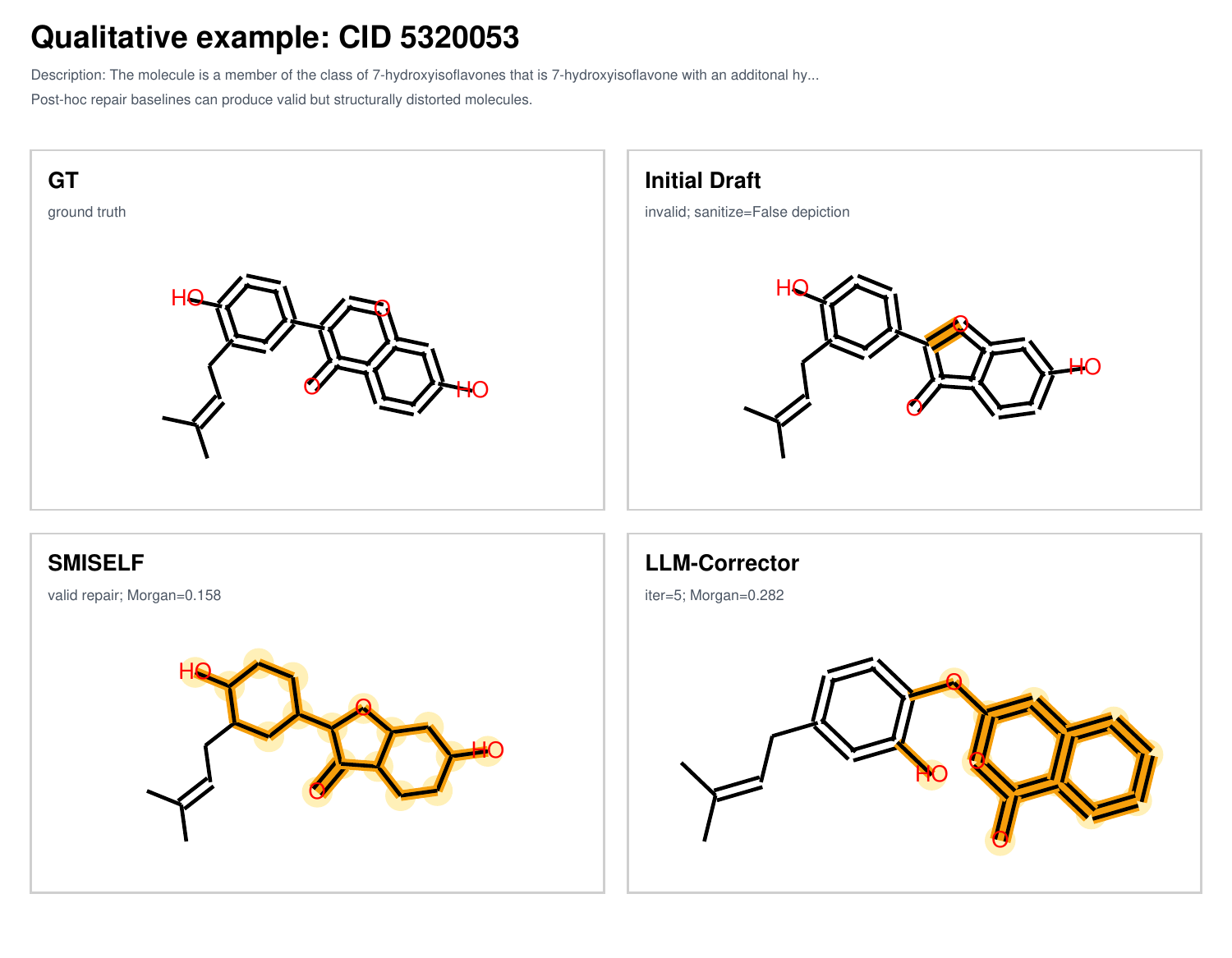} \vspace{-10mm}
    \caption{Qualitative example of molecular restoration for an invalid draft. Although the initial draft preserves target-relevant structural cues, post-hoc repair methods restore validity at the cost of substantial structural drift from the intended molecular identity.}
    \label{fig:qual_case_no_reactt_amrec_5320053} \vspace{-5mm}
\end{figure}

In this section, we analyze why classical correction methods fail to preserve the structural identity of molecular drafts. We then formalize the recovery process as an agentic search over molecular states and identify the key limitations that motivate our proposed framework.

\subsection{The Limits of Current Correction: Why Repair is Not Recovery}

Invalid drafts generated by text-to-molecule models can typically be routed to post-hoc repair pipelines. Rule- or SELFIES-based methods specialize in syntax-level \textit{repair}---patching broken strings into formally valid molecular graphs \citep{weininger1988smiles,krenn2020self,tao2025make}. However, as shown in Table \ref{tab:posthoc_invalid_analysis} and Figure \ref{fig:qual_case_no_reactt_amrec_5320053}, validity-oriented methods can unintentionally alter core scaffolds, functional groups, or other target-relevant substructures that were correctly expressed in the original draft, ultimately reducing semantic alignment with the target description.

A natural alternative is to prompt an LLM to iteratively correct the draft using the target description and execution feedback. Yet this approach remains fundamentally limited. Our empirical analysis (Table \ref{tab:posthoc_invalid_analysis}, Figure \ref{fig:qual_case_no_reactt_amrec_5320053}) shows that LLM-based correction frequently introduces unintended global structural drift, modifying molecular regions that should remain preserved. This limitation arises from the representation itself: LLM-only correction typically regenerates an entire tokenized SMILES sequence at each step, without explicitly separating local edits from globally preserved structures. Thus, attempts to fix one structural issue can inadvertently damage other target-relevant chemical cues.

Consequently, invalid drafts should not be viewed as disposable broken strings to be superficially patched, but rather as \textit{corrupted molecular states} that still preserve meaningful chemical information. Recovering these latent structural cues requires a stateful and context-aware framework that can preserve structural memory, explicitly track molecule-text mismatches, and iteratively validate targeted molecular modifications throughout the recovery process.

\subsection{Molecular Recovery as Agentic Search}

The need for a stateful, iterative correction process naturally motivates an agentic formulation where the molecule itself serves as the environmental state. We formalize this molecular recovery as a sequential decision process over molecular states. Let $\mathcal{M}$ denote the molecular state space and $\mathcal{A}$ represent the action space. Given an initial invalid draft $m_0 \in \mathcal{M}$ and a target natural-language description $d$, the molecular state at step $t$ is denoted by $m_t \in \mathcal{M}$. The action $a_t \in \mathcal{A}$ corresponds to granular molecular operations such as charge alignment, atom/bond editing, or substructure replacement. At each step, the agent selects an action based on the current molecule, target description, validation feedback $o_t$, and interaction history $h_t$:
\begin{equation} \nonumber
        m_{t+1} = \mathcal{T}(m_t, a_t),~a_t \sim \pi_\theta(a \mid m_t, d, o_t, h_t),  
\end{equation}
where $\pi_\theta$ is the agent policy and $\mathcal{T}$ is the transition operator that applies the selected action to the current molecule to produce the next molecular state.

Within this formulation, a key challenge is \textit{action fidelity}: whether an intended local edit is faithfully translated into the actual molecular transition. Integrating executable RDKit-based tools \citep{landrum2013rdkit} partially resolves this issue by grounding molecular actions as deterministic graph operations, providing a tool-grounded transition operator, $\mathcal{T}_{\mathrm{tool}}$, with substantially improved control and reliability. However, improving action fidelity alone is insufficient for robust molecular recovery. While executable tools constrain \textit{how} molecular edits are applied, they do not determine \textit{which} structural requirement should be prioritized first or \textit{whether} alternative recovery strategies should be explored. Consequently, generic tool-augmented agents still operate as a form of \textit{agentic greedy search}, committing to a single locally selected edit at each step while discarding alternative hypotheses.

This greedy formulation exposes two fundamental limitations:
\begin{itemize}[leftmargin=10pt, itemsep=-0.5pt] \vspace{-5pt}
\item \textbf{Alignment Blindness:} Generic agents lack an explicit molecule-aware reasoning mechanism for tracking which structural requirements implied by the target description are already satisfied, still missing, or vulnerable to unintended modification during recovery. \vspace{-3pt}
\item \textbf{Exploration Blindness:} Recovery unfolds along a single linear trajectory, causing the search process to become highly sensitive to early sub-optimal edits that can irreversibly steer the agent toward dead-end recovery paths. \vspace{-5pt}
\end{itemize}
Therefore, effective molecular recovery requires more than faithful executable actions. It demands explicit molecule-text alignment tracking together with broader exploration across multiple candidate trajectories, enabling the agent to preserve target-relevant structural cues while avoiding irreversible recovery failures.

\begin{figure*}[t]
    \centering
    \includegraphics[width=\textwidth]{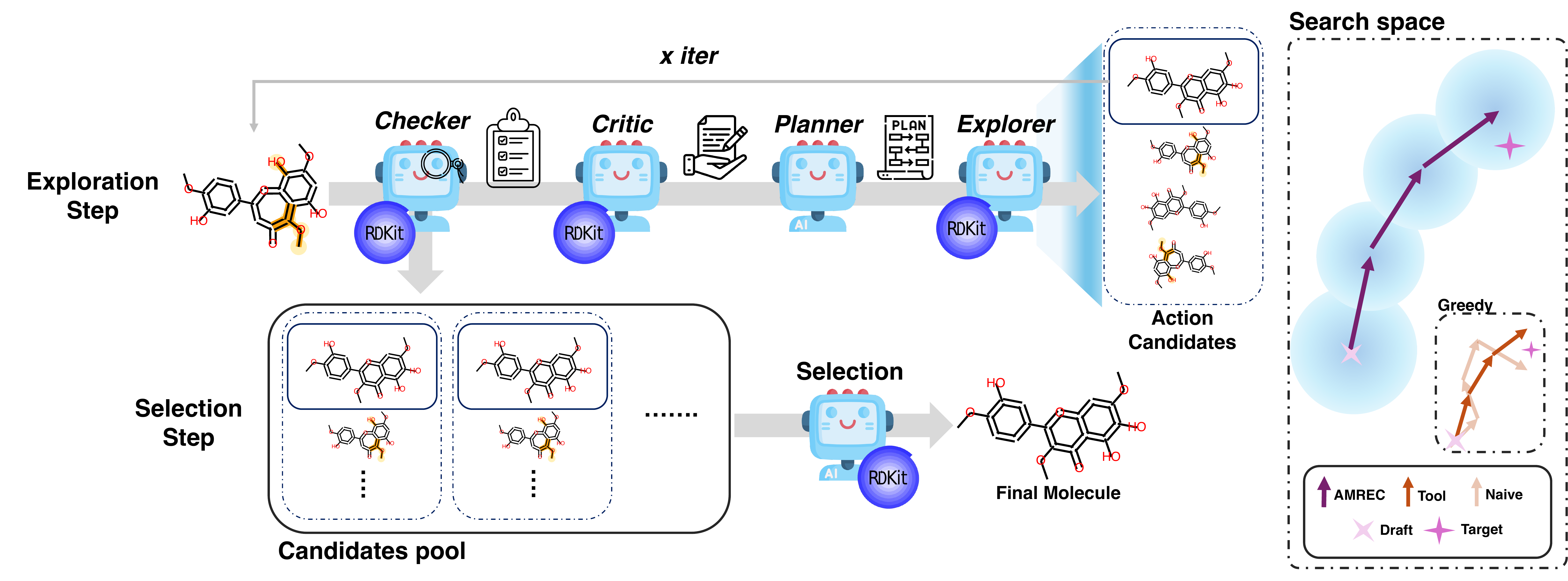} 
    \caption{Overview of \name. \name combines molecular understanding components that track target-property mismatch with expanded search components that construct and revisit a trajectory-level candidate pool.}
    \label{fig:amrec}
            \vspace{-1em}
\end{figure*}

\section{\name: Agentic Molecular Recovery}

To address the limitations of generic tool-augmented agents, \name formulates molecular recovery as a stateful search process grounded in both the target description $d$ and the initial invalid draft $m_0$. Rather than treating the current molecule $m_t$ as the only state variable, \name maintains a richer recovery state at step $t$ by incorporating the interaction history $h_t$ and the cumulative candidate pool $\mathcal{C}_{<t}$ generated up to the previous steps:
\begin{equation}
    s_t = \left(m_t,\ h_t,\ \mathcal{C}_{<t}\right).
\end{equation}
When the initial draft $m_0$ is unparseable, a lightweight bootstrap using SMISELF \citep{tao2025make} is applied as a one-off initialization step, providing a formally valid starting graph from which RDKit properties can be extracted.

On top of this state, \name uses four LLM agents with distinct roles: \texttt{\texttt{Checker}}, \texttt{\texttt{Critic}}, \texttt{\texttt{Planner}}, and \texttt{\texttt{Candidate Explorer}}.
As illustrated in Figure~\ref{fig:amrec}, these modules collaborate along two primary operational axes to resolve the bottlenecks of agentic greedy search:
\begin{itemize}[leftmargin=10pt, itemsep=-0.5pt] \vspace{-5pt}
    \item \textbf{Property-Level Alignment:} \texttt{Checker}, \texttt{Critic}, and \texttt{Planner} form an evaluation-to-action triad that systematically translates natural language into verifiable constraints and targets remaining molecule-text mismatches. \vspace{-3pt}
    \item \textbf{Trajectory-Level Exploration:} \texttt{Candidate Explorer} and a final selection mechanism non-linearly expand the search space, maintaining alternative recovery hypotheses instead of locking onto a singular greedy path.
\end{itemize}


\subsection{\texttt{Checker} Agent}
Before the recovery loop, \name decomposes the target description \(d\) into a fixed set of checkable structural requirements using an LLM: \(\mathcal{P} = \{p_1, p_2, \ldots, p_K\}\). Here, each requirement \(p_i\) corresponds to a molecular property or structural constraint that can be verified against the current molecule. This requirement set serves both as the recovery objective and explicit stopping criterion.

At each iteration, \texttt{Checker} evaluates whether the current molecule $m_t$ satisfies these requirements using RDKit-derived observations:
\begin{align}
    o_t &= \mathrm{RDKit}(m_t), \\
    \mathbf{c}_t &= (c_{t,1},\ldots,c_{t,K}) =  \texttt{\texttt{Checker}}(d,\ m_t,\ \mathcal{P},\ o_t), \nonumber
\end{align}
where \(c_{t,i}=1\) indicates that \(m_t\) satisfies requirement \(p_i\). 
Through this process, \texttt{Checker} converts the high-level molecule-text alignment problem into an explicit property-level checklist. If all requirements are satisfied, recovery terminates immediately, preventing unnecessary modifications that may damage already-correct structures. Unsatisfied requirements are passed to next agents---\texttt{Critic} and \texttt{Planner}---as structured mismatch signals.

\subsection{\texttt{Critic} Agent}
\texttt{Critic} transforms \texttt{Checker}’s property-level outputs into recovery guidance for subsequent planning. It jointly considers the target description $d$, current molecule $m_t$, requirement checklist $\mathcal{P}$, RDKit observations $o_t$, and recovery history $h_t$: 
\begin{equation}
    f_t =
    \mathrm{\texttt{Critic}}(d,\ m_t,\ \mathcal{P},\ \mathbf{c}_t,\ o_t,\ h_t).
\end{equation}
The resulting feedback $f_t$ summarizes unresolved structural mismatches, identifies target-relevant substructures that should be preserved, and highlights potential structural drift introduced in previous steps. \texttt{Critic} therefore acts as a molecule-aware reasoning step that contextualizes the current recovery state and guides future modifications toward unresolved structural objectives while discouraging additional modification that may be unnecessary or harmful.

\subsection{\texttt{Planner} Agent}
\texttt{Planner} converts the \texttt{Critic}'s feedback $f_t$ into an actionable recovery strategy under the current recovery state:
\begin{equation}
    p_t =
    \mathrm{\texttt{Planner}}(d,\ m_0,\ m_t,\ \mathcal{P},\ f_t).
\end{equation}
Here, \(p_t=(\iota_t,\ell_t)\), where \(\iota_t\) denotes the recovery intent that guides the next stage of exploration, and \(\ell_t\) provides a short-horizon rationale describing how the proposed modification may affect scaffold preservation or downstream requirement satisfaction. Guided by this objective, \texttt{Planner} prioritizes minimal structural edits that resolve unsatisfied requirements while avoiding unnecessary changes to target-relevant structural cues, including scaffolds, ring systems, heteroatoms, charge states, stereochemistry, and large substituents. When further modification is likely to introduce harmful structural drift, \texttt{Planner} can explicitly determine that no additional safe modification can be conducted, thereby preventing excessive molecule modification at the planning stage.


\subsection{\texttt{Candidate Explorer} Agent}
\texttt{Candidate Explorer} does not directly execute \texttt{Planner}’s recovery plan \(p_t\) as a single next molecule, but instead expands it into molecular candidates that carry out the same recovery intent in multiple recovery trajectories:
\begin{align}
        \mathcal{C}_t
        &= \{m_{t}^{(1)}, m_{t}^{(2)}, \ldots, m_{t}^{(N)}\} \\
        &= \mathrm{\texttt{Explorer}}(p_t,\ d,\ m_0,\ m_t,\ \mathcal{P},\ f_t, \ o_t,\ h_t). \nonumber
\end{align}
\texttt{Candidate Explorer} conditions jointly on the target description, initial draft, current molecule, planning intent, recovery history, and structural feedback. As a result, $\mathcal{C}_t$ represents a structured set of recovery hypotheses rather than a single one-shot proposal. This allows multiple draft-preserving and description-consistent recovery directions to be explored simultaneously within the same iteration.

From $\mathcal{C}_t$, \name temporarily selects one valid candidate as the provisional next state $m_{t+1} \in \mathcal{C}_t$, which is subsequently re-evaluated by \texttt{Checker} and \texttt{Critic}. Importantly, unselected candidates are not discarded. Instead, all explored candidates are preserved in the recovery history and reconsidered during final trajectory-level selection. Unlike conventional \textit{agentic greedy search}, which commits exclusively to the latest state, \name continuously expands and preserves a broader candidate pool throughout the recovery process.



\subsection{Trajectory-level Candidate Selection}
The recovery loop iteratively repeats the \texttt{Checker}--\texttt{Critic}--\texttt{Planner}--\texttt{Candidate Explorer} cycle until either all structural requirements are satisfied or the maximum iteration budget is reached. Conventional greedy-search agents would typically return the final molecular state as the output. However, in molecular recovery, later modifications may inadvertently damage previously preserved structural cues, meaning earlier candidates can sometimes better preserve the intended molecular identity.

To address this issue, \name performs trajectory-level candidate selection after the recovery loop terminates. First, it aggregates all explored candidates into a unified candidate pool:
\begin{equation}
    \mathcal{C}_{\mathrm{all}}
    =
    \mathrm{\texttt{Collect}}
    \big(
    m_T,\ h_T,\ \{\mathcal{C}_t\}_{t=0}^{T-1}
    \big).
\end{equation}
Here, \(\mathcal{C}_{\mathrm{all}}\) includes the final state, intermediate candidates generated during recovery, and auxiliary candidates preserved during initialization. 
\texttt{Final Selector} then chooses one molecule from this trajectory-level candidate pool \(\mathcal{C}_{\mathrm{all}}\):
\begin{equation}
    \hat{m}
    =
    \mathrm{\texttt{Select}}(d,\ m_0,\ \mathcal{C}_{\mathrm{all}}).
\end{equation}

Rather than generating a new molecule, \texttt{Select} performs comparative evaluation across the entire recovery trajectory, balancing target satisfaction against preservation of target-relevant structural cues from the original draft. In this way, \name avoids overcommitting to a single greedy trajectory and instead reframes molecular recovery as an explicitly stateful, molecule-aware, and trajectory-level search process.

\section{Experiment}
\begin{table*}[t]
\centering
\scriptsize
\setlength{\tabcolsep}{7pt}
\renewcommand{\arraystretch}{0.9}
\caption{Main correction results at ChEBI-20 dataset grouped by backbone model. T denotes tool.} \vspace{-2mm}
\label{tab:main_results}
\begin{tabular}{l|c|c|c|c|c|c|c|c}
\toprule
\rowcolor{gray!18}
Method & MACCS FTS$\uparrow$ & RDK FTS$\uparrow$ & Morgan FTS$\uparrow$ & Exact$\uparrow$ & BLEU$\uparrow$ & ROUGE-L$\uparrow$ & Levenshtein$\downarrow$ & FCD$\downarrow$ \\
\midrule
\rowcolor{gray!10}
\multicolumn{9}{c}{\textit{GPT-5.4-mini}} \\
Initial Raw & -- & -- & -- & -- & 0.6407 & 0.7247 & 46.4433 & -- \\
SMISELF & 0.7154 & 0.3708 & 0.2628 & 0.0000 & 0.4467 & 0.6295 & 50.8814 & 18.4483 \\
LLM-Corrector & 0.7505 & 0.4407 & 0.3239 & 0.0567 & 0.4979 & 0.6547 & 49.9948 & 14.7324 \\
\midrule
ReAct & 0.7766 & 0.4873 & 0.3823 & 0.0515 & 0.5574 & 0.6954 & 41.7938 & 13.0870 \\
ReWOO & \underline{0.8274} & 0.5881 & 0.4774 & \underline{0.1495} & 0.6207 & 0.7427 & 36.7887 & 11.3297 \\
PlanAndAct & 0.7770 & 0.4882 & 0.3802 & 0.0619 & 0.5516 & 0.6945 & 48.3711 & 12.6822 \\
\midrule
ReAct-T & 0.8079 & \underline{0.6276} & \underline{0.4915} & 0.0979 & \underline{0.6383} & \underline{0.7489} & 36.7216 & 8.8410 \\
ReWOO-T & 0.8152 & 0.5672 & 0.4509 & 0.1134 & 0.6214 & 0.7383 & \underline{35.8351} & 10.3550 \\
PlanAndAct-T & 0.8082 & 0.6079 & 0.4789 & 0.0773 & 0.6173 & 0.7397 & 39.9021 & \underline{8.4538} \\
\midrule
\rowcolor{blue!10}
\name & \textbf{0.8661} & \textbf{0.6796} & \textbf{0.5760} & \textbf{0.1959} & \textbf{0.7166} & \textbf{0.8007} & \textbf{31.2526} & \textbf{7.4510} \\
\midrule
\rowcolor{gray!10}
\multicolumn{9}{c}{\textit{Gemini-3.1-Flash-Lite}} \\
Initial Raw & -- & -- & -- & -- & 0.6357 & 0.7555 & 25.2214 & -- \\
SMISELF & 0.6932 & 0.3493 & 0.2729 & 0.0214 & 0.4247 & 0.6172 & 34.1000 & 18.3432 \\
LLM-Corrector & 0.7902 & 0.4906 & 0.4164 & 0.1214 & 0.5658 & 0.7058 & 34.7214 & 12.9917 \\
\midrule
ReAct & 0.7955 & 0.5193 & 0.4472 & 0.1714 & 0.5841 & 0.7161 & 29.5500 & 10.7484 \\
ReWOO & 0.8442 & 0.6103 & 0.5368 & \underline{0.2357} & 0.6543 & 0.7570 & 27.2500 & 8.8451 \\
PlanAndAct & 0.7918 & 0.5275 & 0.4414 & 0.1143 & 0.5891 & 0.7305 & 30.4786 & 10.5412 \\
\midrule
ReAct-T & \underline{0.8443} & \textbf{0.6480} & \underline{0.5594} & 0.1929 & \underline{0.6842} & \underline{0.7880} & \underline{23.8500} & \textbf{7.3022} \\
ReWOO-T & 0.8402 & 0.5950 & 0.5379 & 0.2214 & 0.6477 & 0.7550 & 27.7143 & 8.7526 \\
PlanAndAct-T & 0.8041 & 0.6009 & 0.5083 & 0.1286 & 0.6437 & 0.7612 & 25.0929 & \underline{8.3742} \\
\midrule
\rowcolor{blue!10}
\name & \textbf{0.8618} & \underline{0.6322} & \textbf{0.5745} & \textbf{0.2429} & \textbf{0.6988} & \textbf{0.7915} & \textbf{23.5214} & 9.8556 \\
\midrule
\rowcolor{gray!10}
\multicolumn{9}{c}{\textit{Claude-haiku-4.5}} \\
Initial Raw & -- & -- & -- & -- & 0.6328 & 0.7439 & 24.5405 & -- \\
SMISELF & 0.6920 & 0.3788 & 0.2760 & 0.0203 & 0.4326 & 0.6226 & 29.7703 & 13.7593 \\
LLM-Corrector & 0.5704 & 0.3662 & 0.2897 & 0.0676 & 0.3913 & 0.5213 & 38.3784 & 18.8593 \\
\midrule
ReAct & 0.7886 & 0.5215 & 0.4346 & 0.1014 & 0.5902 & 0.7202 & 24.8649 & 7.7959 \\
ReWOO & 0.8085 & 0.5337 & 0.4514 & 0.1351 & 0.6173 & 0.7427 & 23.8243 & 7.2375 \\
PlanAndAct & 0.7607 & 0.4626 & 0.3830 & 0.0608 & 0.5441 & 0.6987 & 25.7297 & 9.0560 \\
\midrule
ReAct-T & 0.8088 & 0.5758 & 0.4814 & 0.1014 & 0.6428 & 0.7544 & 22.5473 & 5.9749 \\
ReWOO-T & \underline{0.8362} & 0.5894 & \underline{0.4959} & \underline{0.1419} & \underline{0.6619} & \underline{0.7576} & \underline{22.3514} & \underline{5.6864} \\
PlanAndAct-T & 0.8096 & \underline{0.5942} & 0.4863 & 0.0946 & 0.6479 & 0.7537 & 22.5000 & 5.7312 \\
\midrule
\rowcolor{blue!10}
\name & \textbf{0.8577} & \textbf{0.6347} & \textbf{0.5408} & \textbf{0.1689} & \textbf{0.7088} & \textbf{0.7917} & \textbf{20.3243} & \textbf{5.1682} \\
\bottomrule
\end{tabular}%
\end{table*}

\begin{figure*}[t]
    \centering
    \includegraphics[width=\textwidth]{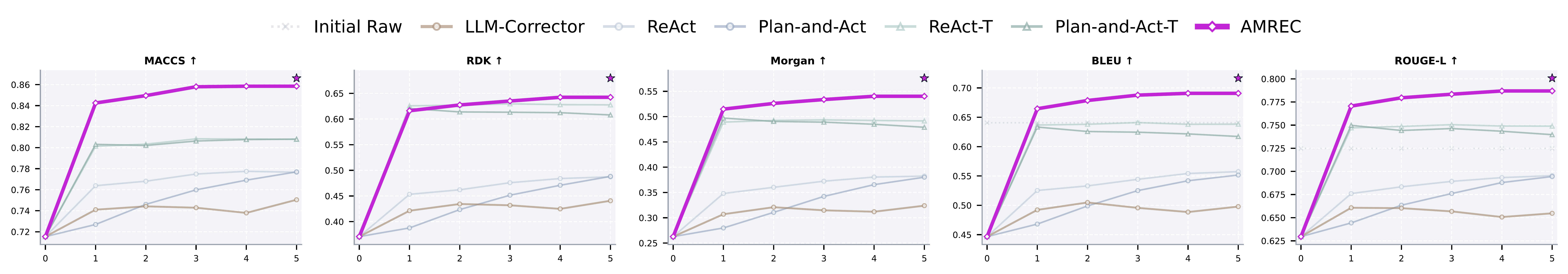} 
    \vspace{-2em}
    \caption{Intermediate recovery behavior on GPT-5.4-mini. The curves follow each method's native execution stages under the same invalid initial subset. Purple stars indicate results after trajectory-level candidate selection.} 
    \label{fig:main_methods_iteration_all_metrics}
        \vspace{-1em}
\end{figure*}
\begin{figure*}[t]
    \centering
    \begin{minipage}[t]{0.48\textwidth}
        \centering
        \includegraphics[width=\linewidth]{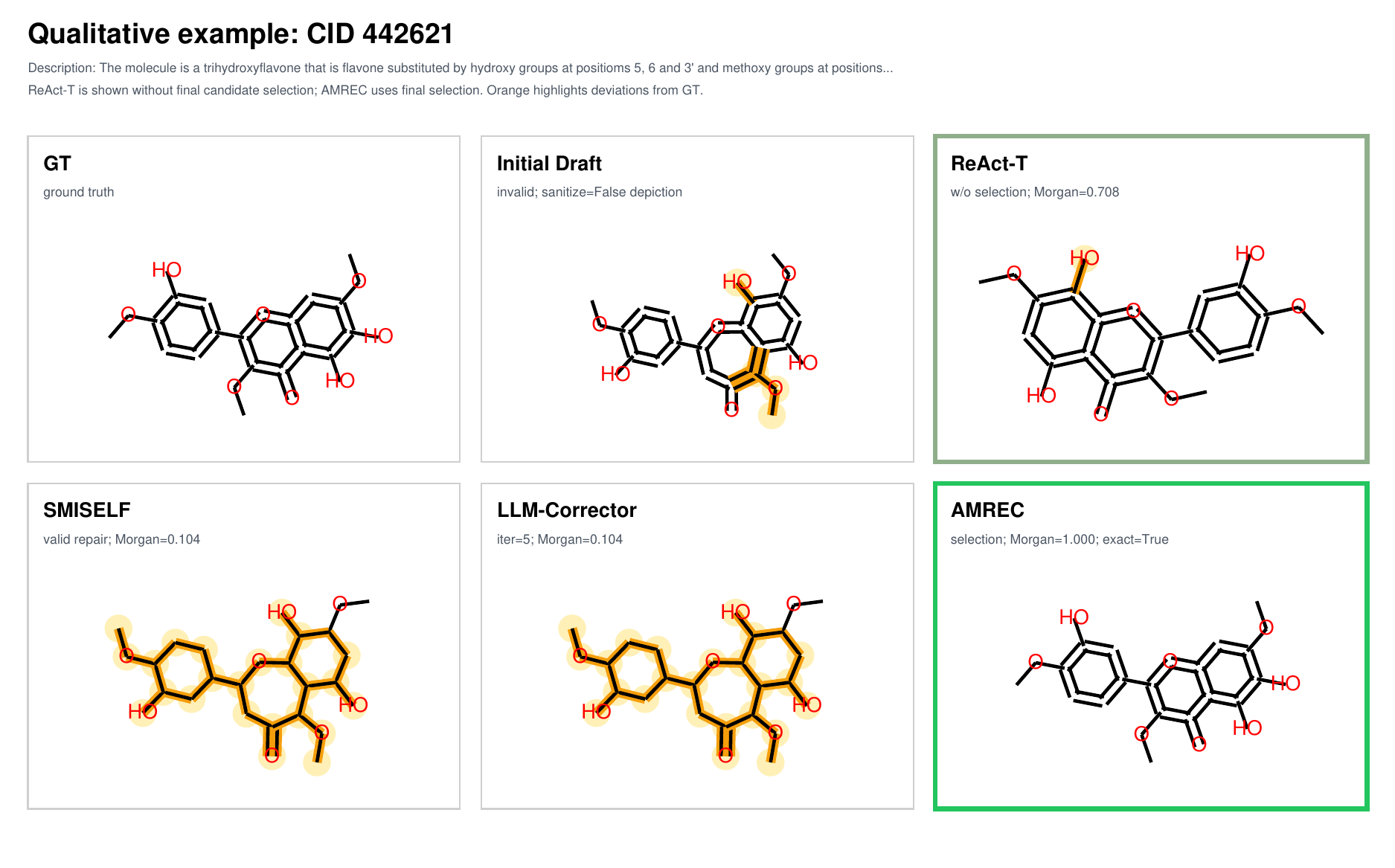}
    \end{minipage}
    \hfill
    \begin{minipage}[t]{0.48\textwidth}
        \centering
        \includegraphics[width=\linewidth]{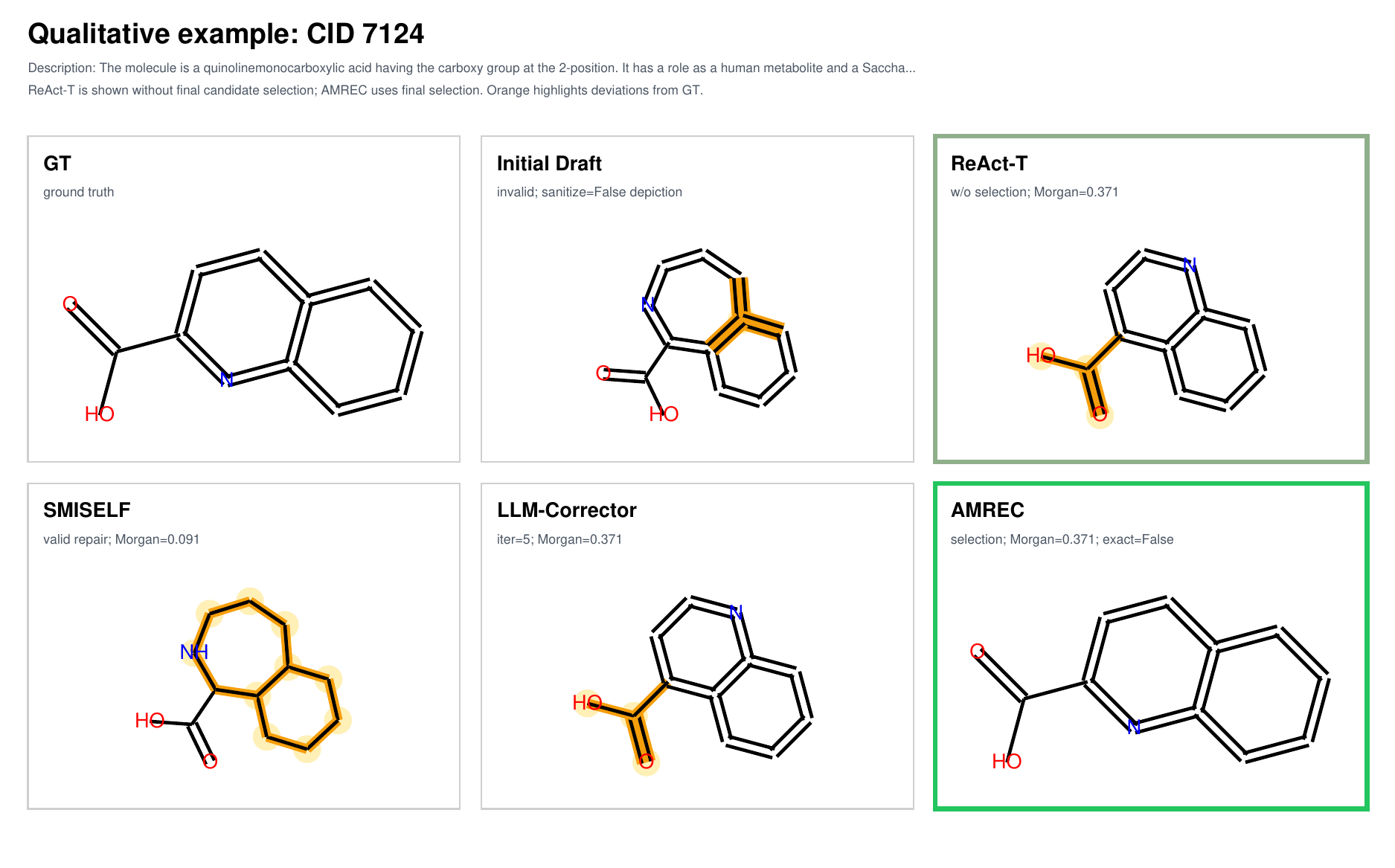}
    \end{minipage} \vspace{-5mm}
    \caption{Qualitative comparison between tool-augmented generic agents and \name. \name recovers target-relevant molecular structure more faithfully, while greedy edits can still leave residual errors or introduce new distortions.} \vspace{-3mm}
    \label{fig:qualitative_cases_amrec_react_t}
\end{figure*}


In this section, we evaluate the effectiveness of \name for molecular recovery from invalid drafts. In particular, we compare \name against validity-oriented repair methods, LLM-only correction, and generic tool-grounded agentic search to examine how molecule-aware mismatch tracking and trajectory-level candidate exploration contribute to preserving target-relevant structural cues during molecular restoration.

\subsection{Experimental Setup}
We evaluate molecular recovery on invalid initial drafts generated from the ChEBI-20 validation split \citep{edwards2021text2mol}.
All methods are given the same target description $d$ and initial draft \(m_0\), and the final output is compared against the ground-truth target molecule. We use GPT-5.4-mini, Gemini-3.1-flash-lite and Claude-haiku-4.5 as backbone models.

\define{Baselines} We compare validity-oriented repair, LLM-only correction, greedy agentic recovery, tool-augmented greedy recovery, and \name.
\textbf{Initial Raw} is the original invalid draft, \textbf{SMISELF} performs post-hoc syntax-level repair \citep{tao2025make}, and \textbf{LLM-Corrector} iteratively rewrites the molecule using the target description and validity feedback.
For generic agentic baselines, we use ReAct, ReWOO, and PlanAndAct with RDKit observations \citep{yao2022react,xu2023rewoo,erdogan2025plan, landrum2013rdkit}.
Their tool-augmented variants, ReAct-T, ReWOO-T, and PlanAndAct-T, additionally use executable molecular edit tools in Table~\ref{tab:rdkit_edit_actions}. For all agentic methods, SMISELF is applied initially to use RDkit edit and its observation.
ReAct- and PlanAndAct-style baselines are run for five iterations.
For \name, we use five candidates per iteration and a maximum of five iterations, where the last stage performs trajectory-level candidate selection.
Any remaining invalid outputs are repaired using SMISELF as a fallback for fair evaluation, so all molecules reported in the tables are valid. The detailed experimental setup is provided in Appendix~\ref{Setup}.

\define{Metrics} We use metrics covering structural similarity(MACCS, RDK, Morgan FTS), exact identity recovery(Exact), string-level similarity(BLEU, ROUGE-L, Levenshtein), and distribution-level distance(FCD); a full list and detailed descriptions are provided in Appendix~\ref{app:metrics}.

\subsection{Main Results}
Table~\ref{tab:main_results} reports the recovery results from invalid drafts under three backbone models. Overall, \name achieves the best performance across all metrics, showing the strongest structural, exact-match, string-level, and distribution-level recovery.

Generic agentic methods outperform LLM-only correction by making recovery iterative and by using RDKit-derived factual information.
However, their gains remain limited because they still refine a single accepted molecule at each step, making later recovery vulnerable to early structural drift.

Tool-augmented variants generally improve over their non-tool counterparts.
Nevertheless, the gains remain limited because these agents still follow greedy single-candidate refinement and lack molecule-aware mismatch tracking between the target description and the current molecular state.
For agents with a more structured expansion procedure, such as ReWOO, the tool-based constraint can also narrow the explored space and occasionally lower performance.
These results suggest that robust recovery requires molecular understanding and expanded candidate search.

The improvement of \name comes from addressing this remaining limitation. \name uses description-derived requirements and \texttt{Checker}--\texttt{Critic}--\texttt{Planner} guidance to track target-relevant mismatches, and then uses \texttt{Candidate Explorer} and trajectory-level selection to move beyond tool-grounded greedy recovery. In particular, candidate exploration improves the fidelity with which a planned recovery intent is realized as a molecular transition: instead of realizing one intent as a single next molecule, \name generates multiple candidate realizations and compares them with respect to target satisfaction and draft preservation. Therefore, the gain of \name reflects not only safer local edit execution, but also expanded candidate search that reduces premature greedy commitment.

Figure~\ref{fig:main_methods_iteration_all_metrics} further supports this interpretation. Tool-augmented generic agents improve after the first executable edit, but their trajectories still follow a single accepted path; consequently, several metrics decline as iterations proceed. In contrast, \name moves toward more target-relevant candidates from early stages and continues to refine remaining structural details through candidate exploration. Trajectory-level selection also allows \name to select the best molecule from the explored trajectory, rather than forcing the final output to be the last state of the loop.

Figure~\ref{fig:qualitative_cases_amrec_react_t} qualitatively illustrates the same distinction. Tool-augmented generic agents can perform local edits, but they may still miss target-relevant cues or distort other substructures because they remain greedy. \name instead expands the candidate set and compares molecules across the trajectory, enabling more faithful recovery of the target scaffold and key substructures. Thus, the performance gap in Table~\ref{tab:main_results} reflects more than numerical improvement; it shows that invalid molecule recovery requires going beyond valid repair and tool-grounded greedy editing toward molecule-aware candidate exploration.

\section{Ablation Studies}

\define{Effect of candidate pool expansion}
Table~\ref{tab:amrec_prefix_all_metrics_iter1_4} shows that using a larger trajectory-level pool leads to better recovery quality.
The improvement is most visible in structure-sensitive and identity-level metrics, indicating that useful candidates can emerge at different stages of the recovery loop.
This supports our design choice of retaining intermediate candidates instead of relying only on the terminal molecule.

\begin{table}[t]
\centering
\scriptsize
\setlength{\tabcolsep}{4pt}
\renewcommand{\arraystretch}{1.05}
\caption{
Effect of trajectory-level candidate pool expansion in \name on GPT-5.4-mini.
Iter \(k\) uses candidates accumulated up to iteration \(k\) for final selection.
} \vspace{-2mm}
\label{tab:amrec_prefix_all_metrics_iter1_4}
\begin{tabular}{l|c|c|c|c}
\toprule
\rowcolor{gray!18}
Metric & Iter 1 & Iter 2 & Iter 3 & Iter 4 \\
\midrule
MACCS$\uparrow$ & 0.8508 & 0.8548 & 0.8569 & \textbf{0.8661} \\
RDK$\uparrow$ & 0.6589 & 0.6588 & 0.6648 & \textbf{0.6796} \\
Morgan$\uparrow$ & 0.5499 & 0.5533 & 0.5547 & \textbf{0.5760} \\
Exact$\uparrow$ & 0.1392 & 0.1443 & 0.1443 & \textbf{0.1959} \\
BLEU$\uparrow$ & 0.6937 & 0.6977 & 0.7014 & \textbf{0.7166} \\
ROUGE-L$\uparrow$ & 0.7848 & 0.7855 & 0.7850 & \textbf{0.8007} \\
Levenshtein$\downarrow$ & 32.8454 & 32.9794 & 33.1701 & \textbf{31.2526} \\
FCD$\downarrow$ & 8.3462 & 8.2768 & 8.3033 & \textbf{7.4510} \\
\bottomrule
\end{tabular}
\end{table}

\define{Effect of \texttt{Critic}}
Table~\ref{tab:critic} shows that \texttt{Critic} contributes to more accurate structural recovery.
By converting checker outputs into targeted feedback, it helps \texttt{Planner} prioritize remaining mismatches while reducing unnecessary edits.
This suggests that molecule-aware feedback is important for guiding exploration beyond simple iterative correction.

\begin{table}[t]
\centering
\scriptsize
\setlength{\tabcolsep}{3pt}
\renewcommand{\arraystretch}{0.9}
\caption{Effect of Critic on GPT-5.4-mini. Each entry shows without critic → with critic.} \vspace{-2mm}
\label{tab:critic}
\begin{tabular}{l|c|c|c}
\toprule
\rowcolor{gray!18}
Method & MACCS FTS$\uparrow$ & RDK FTS$\uparrow$ & Morgan FTS$\uparrow$ \\
\midrule
\rowcolor{blue!10}
\name & 0.8456 $\rightarrow$ \textbf{0.8661} & 0.6419 $\rightarrow$ \textbf{0.6796} & 0.5200 $\rightarrow$ \textbf{0.5760} \\
\bottomrule
\end{tabular}%
\end{table}

\define{Effect of final candidate selection}
Table~\ref{tab:baseline_t_selection_comparison} shows that final selection improves recovery by choosing from the explored trajectory rather than taking the last output.
The gains also appear in tool-augmented baselines, confirming that terminal states are not always optimal.
However, \name remains strongest after selection, suggesting that selection is most effective when the preceding trajectory contains high-quality candidates.
\begin{table}[t]
\centering
\scriptsize
\setlength{\tabcolsep}{3pt}
\renewcommand{\arraystretch}{0.9}
\caption{
Effect of final candidate selection on GPT-5.4-mini.
Each entry shows terminal output \(\rightarrow\) selected output from the candidate pool.
} \vspace{-2mm}
\label{tab:baseline_t_selection_comparison}
\begin{tabular}{l|c|c|c}
\toprule
\rowcolor{gray!18}
Method & MACCS FTS$\uparrow$ & RDK FTS$\uparrow$ & Morgan FTS$\uparrow$ \\
\midrule
ReAct-T & 0.8079 $\rightarrow$ 0.8160 & \underline{0.6276} $\rightarrow$ \underline{0.6412} & \underline{0.4915} $\rightarrow$ \underline{0.5145} \\
ReWOO-T & \underline{0.8152} $\rightarrow$ \underline{0.8307} & 0.5672 $\rightarrow$ 0.6340 & 0.4509 $\rightarrow$ 0.5090 \\
PlanAndAct-T & 0.8082 $\rightarrow$ 0.8153 & 0.6079 $\rightarrow$ 0.6228 & 0.4789 $\rightarrow$ 0.4987 \\
\midrule
\rowcolor{blue!10}
\name & \textbf{0.8585} $\rightarrow$ \textbf{0.8661} & \textbf{0.6424} $\rightarrow$ \textbf{0.6796} & \textbf{0.5402} $\rightarrow$ \textbf{0.5760} \\
\bottomrule
\end{tabular}%
\end{table}

\define{Efficiency of \name}
Table~\ref{tab:computation_analysis} shows that \name does not require exhausting the full iteration budget in most cases.
This indicates that the recovery loop can often identify sufficient candidates early, while still preserving additional iterations for more difficult cases.
Thus, \name improves recovery quality with a modest iterative cost.
\begin{table}[t]
\centering
\scriptsize
\setlength{\tabcolsep}{5pt}
\renewcommand{\arraystretch}{1.05}
\caption{Computation analysis of \name. Each iteration column reports the number of samples terminated at that iteration; Mean denotes the average number of executor iterations.}
\label{tab:computation_analysis}
\begin{tabular}{l|c|c|c|c|c}
\toprule
\rowcolor{gray!18}
Model & Iter 1 & Iter 2 & Iter 3 & Iter 4 & Mean \\
\midrule
GPT-5.4-mini & 130  & 22  & 9  & 33  & 1.72 \\
Gemini-3.1-Flash-Lite & 117  & 6  & 1  & 16  & 1.40 \\
Claude-haiku-4.5 & 120  & 8  & 4  & 16  & 1.43 \\
\bottomrule
\end{tabular}
\end{table}



\section{Conclusion}
We studied invalid molecular outputs as corrupted molecular states that require recovery rather than simple syntax repair. 
Our experiments show that post-hoc repair and LLM-only correction do not reliably recover target-relevant molecular identity. 
Executable molecular edit tools improve recovery performance, but their gains remain limited by greedy single-candidate search. We proposed \name to address this limitation through molecule-aware mismatch tracking, candidate exploration, and trajectory-level selection. 
The results demonstrate that recovery is more effective when the method expands and compares candidate trajectories instead of committing to the final state of a single greedy path.


\newpage
\section*{Limitations}

This work focuses on computational recovery of invalid SMILES outputs
under benchmark settings. The recovered molecules are evaluated using
automatic structural, exact-match, string-level, and distribution-level
metrics, which do not establish synthesizability, biological activity,
toxicity, or real-world utility. In addition, AMREC relies on LLM-based
requirement extraction, criticism, planning, and candidate selection, and
its behavior may vary across backbone models, prompts, and chemical
domains. Future work should evaluate broader molecular datasets,
stronger chemical validation protocols, and expert-reviewed safety
screening.

\section*{Ethical Considerations}

This work is a computational study of invalid SMILES recovery for
text-guided molecular generation. It uses benchmark molecular-description
data and does not involve human participants, personally identifiable
information, private user data, animal subjects, wet-lab experiments, or
clinical deployment. The proposed method is intended for benchmarking
molecular representation recovery rather than for automated drug
discovery, synthesis planning, or chemical safety assessment.

A potential risk is that improved recovery of valid molecular structures
could be misused when paired with harmful target descriptions or
downstream synthesis and optimization tools. Recovered molecules may
include chemically plausible structures, but they should not be
interpreted as safe, synthesizable, biologically active, or suitable for
real-world use without expert review and independent validation. Any
deployment should include safeguards such as expert oversight, screening
for hazardous or controlled compounds, restrictions on high-risk prompts,
and provenance logging.

\label{sec:bibtex}


\bibliography{custom}

\appendix
\newpage
\newpage

\setcounter{figure}{0}
\setcounter{table}{0}

\paragraph{The Use Of Large Language Model Assistants}
We drafted the manuscript ourselves, and used ChatGPT-5.5 and ChatGPT-5.5-Pro.
\section{Additional Related Work}~\label{Related}
\define{Remark on terminology}
We slightly abuse the terminology of syntactic and semantic errors in this paper.
In SMISELF~\citep{tao2025make}, syntactic errors refer to invalid SMILES strings
that cannot be parsed into molecular graphs, whereas semantic errors refer to
parsed molecular graphs that violate basic chemical constraints, such as atom
valence rules. In our introduction, when we refer to \emph{syntax errors} in
invalid molecular drafts, we use the term broadly to cover both syntactic and
single-molecule-level semantic errors in the sense of SMISELF. By contrast, our
use of \emph{semantic error} is reserved for a different level of mismatch: the
contextual misalignment between the target natural-language description and the
recovered molecule. Thus, semantic recovery in our setting does not merely mean
obtaining a chemically valid molecular graph, but recovering a molecule whose
scaffold, functional groups, and other target-relevant structural cues are
consistent with the description.

\paragraph{Text-guided molecular de novo generation.}
Text-guided molecular de novo generation aims to generate molecular structures from natural-language descriptions of desired chemical structures, functions, or properties. This setting has been studied through molecule-text retrieval, molecule captioning, and text-based molecule generation with task-specific models, pretrained molecule-language models, and large language models, including Text2Mol, MolT5, MolXPT, BioT5, MoleculeSTM, MolReGPT, and recent diffusion-based variants such as TGM-DLM and TextSMOG~\citep{edwards2021text2mol,edwards2022translation,liu2023molxpt,pei2023biot5,liu2023moleculestm,li2024empowering,gong2024text,luo2024textguideddiffusionmodel3d}. These approaches demonstrate that natural language can provide a flexible interface for molecular design, but they mainly focus on generating molecules directly from text. In contrast, our work studies post-generation molecular recovery, where an invalid draft may still contain target-relevant structural cues that should be preserved.

\paragraph{Validity-oriented molecular representations and repair.}
A central challenge in SMILES-based molecular generation is validity: small errors in branches, rings, aromaticity, or valence can make a generated string unparsable~\citep{weininger1988smiles}. Prior work has addressed this issue using validity-preserving or grammar-aware molecular representations, including SELFIES, DeepSMILES, GrammarVAE, Syntax-Directed VAE, Group SELFIES, and related robust molecular languages~\citep{krenn2020self,oboyle2018deepsmiles,kusner2017grammar,dai2018syntax,cheng2023group}. Other methods instead perform post-hoc repair or correction of invalid molecular strings, such as UnCorrupt SMILES and SMISELF~\citep{schoenmaker2023uncorrupt,tao2025make}. These methods are effective for restoring chemical validity, but validity alone does not ensure that the repaired molecule preserves the structural identity implied by the target text. Our work therefore distinguishes syntactic repair from molecular recovery.

\paragraph{Molecular optimization.}
Molecular optimization modifies existing molecules to improve target properties while often maintaining structural similarity to a starting molecule. Classical approaches formulate this process over molecular graphs, latent spaces, or sequential decision processes, including JT-VAE, GCPN, MolDQN, HierVAE, GraphAF, and related graph-based generative models~\citep{Jin2018JunctionTV,You2018GraphCP,Zhou2018OptimizationOM,Jin2020HierarchicalGO,Shi2020GraphAFAF}. More recent approaches incorporate language models or prompting into molecular optimization, such as Prompt-MolOpt, DrugAssist, AgentDrug, LICO, and ReMol~\citep{Wu2024LeveragingLM,Ye2023DrugAssistAL,le2024agentdrug,Nguyen2024LICOLL,WangReMolLM}. These methods are related to our iterative editing setting, but they are typically property-driven: the goal is to improve a valid molecule with respect to desired objectives rather than recover the intended molecular identity from an invalid draft.

\paragraph{Chemical agents.}
LLM-based agents combine reasoning, planning, tool use, and iterative feedback for complex task solving~\citep{yao2022react,xu2023rewoo,erdogan2025plan}. In chemistry, agentic frameworks have been applied to chemical reasoning, synthesis planning, property evaluation, inverse design, experiment automation, and drug discovery, including ChemCrow, Coscientist, ChemAgent, ChemToolAgent, dziner, AgentDrug, MT-MOL, and ToolMol~\citep{m2024augmenting,boiko2023autonomous,Tang2025ChemAgentSL,yu2025tooling,ansari2024dziner,le2024agentdrug,Kim2025MTMolMultiAS,zhou2026toolmol}. These works show the promise of tool-augmented chemical reasoning, especially when LLM decisions are grounded by executable chemistry tools such as RDKit~\citep{landrum2013rdkit}. However, existing molecule-level agents are mainly designed for task completion or property-driven optimization, and do not explicitly address identity-preserving recovery from invalid molecular drafts.

\renewcommand{\thefigure}{B.\arabic{figure}}
\renewcommand{\thetable}{B.\arabic{table}}
\section{Experiment Setup: Detailed}~\label{Setup}
\paragraph{Backbones.}
We use three backbone models, 
\begin{itemize}[leftmargin=1.0em]
\item \texttt{openai/gpt-5.4-mini-20260317}
\item \texttt{google/gemini-3.1-flash-lite-20260507}
\item \texttt{anthropic/claude-4.5-haiku-20251001}
\end{itemize}
accessed through
OpenRouter. We set the decoding temperature to 0 for all baselines and agentic modules. For the \texttt{Candidate Explorer} in \name, we use a temperature of 0.5 to encourage diverse candidate generation during trajectory-level exploration.
For the main-table experiments with single run, we additionally estimate an upper-bound end-to-end inference budget under the worst-case assumption that \name uses the maximum number of iterations for every sample. Under this conservative setting, one backbone-model experiment requires 34.38M tokens in total, corresponding to an estimated output cost of \$51.57 for Gemini-3.1-Flash-Lite under OpenRouter pricing. This estimate should be interpreted as a worst-case upper bound: in practice, \name terminates much earlier on average, requiring only 1.4 iterations for Gemini-3.1-Flash-Lite, and therefore the average inference cost is substantially lower.
\begin{table}[h]
\centering
\scriptsize
\begin{tabular}{l|c}
\hline
\textbf{Model} & \textbf{invalid} \\
\hline
GPT-5.4-mini (\texttt{openai/gpt-5.4-mini}) & 194 \\
Claude-haiku-4.5 (\texttt{anthropic/claude-haiku-4.5}) & 148 \\
Gemini-3.1-Flash-Lite (\texttt{google/gemini-3.1-flash-lite}) & 140 \\
\hline
\end{tabular}
\caption{Number of invalid molecules}
\label{tab:model_comparison}
\end{table}

\paragraph{Dataset.}
In our setting, the goal is not to generate molecules
from scratch, but to recover chemically valid and target-consistent molecules from invalid
initial drafts.

We use the publicly available ChEBI-20-MM benchmark from Hugging Face solely for research purposes \citep{edwards2021text2mol, liu2025quantitative}. Our experiments use only the description–SMILES pairs, and we follow the dataset’s associated access conditions and license terms. ChEBI-20-MM dataset is a text-to-molecule benchmark
consisting of natural-language molecular descriptions paired with ground-truth molecular
structures represented as SMILES. Specifically, we use the ChEBI-20-MM validation split and first generate initial molecular drafts
using three backbone text-guided generation models: GPT-5.4-mini,
Gemini-3.1-Flash-Lite, and Claude-haiku-4.5. We then construct the evaluation subset
from test examples for which the corresponding backbone model produces an invalid
SMILES. All recovery methods are evaluated on these invalid initial drafts under the same
target description and ground-truth molecule. 
The number of invalid initial drafts produced by each backbone model is reported in Table~\ref{tab:model_comparison}.

\paragraph{Evaluation Metrics.} \label{app:metrics}
We evaluate recovery quality using structural, exact-match, string-level, and
distribution-level metrics. For structural similarity, we report Tanimoto similarity based on MACCS, RDKit, and Morgan fingerprints~\citep{bajusz2015tanimoto, durant2002reoptimization, schneider2015get, rogers2010extended}. These metrics compare the recovered molecule
with the ground-truth molecule under different molecular fingerprint representations, and
higher values indicate better structural agreement.

Exact Match measures whether the recovered molecule is identical to the ground-truth
molecule. BLEU and ROUGE-L evaluate string-level overlap between the recovered and
reference molecular representations, while Levenshtein distance measures the minimum
number of edits required to transform the recovered string into the reference string~\citep{papineni2002bleu, lin2004rouge,levenshtein1966binary}. Higher
BLEU and ROUGE-L scores are better, whereas lower Levenshtein distance is better.
Finally, FCD measures the distributional distance between recovered and reference
molecules in the ChemNet embedding space, where lower values indicate better
distributional alignment~\citep{mayr2018large}.

\paragraph{Packages and hyperparameters.}
We used RDKit 2026.03.2 for SMILES validity checking, canonicalization, InChI exact match, and fingerprint computation. Fingerprint Tanimoto scores used RDKit defaults for MACCS keys and RDK fingerprints, and Morgan fingerprints with radius 2. FCD was computed with the \texttt{fcd} package v1.2.2 using its default pretrained model. BLEU was computed with NLTK v3.9.4 \texttt{corpus\_bleu}; Levenshtein distance used \texttt{python-Levenshtein} v0.27.3; ROUGE-L was our character-level LCS implementation. BM25 retrieval used \texttt{rank\_bm25} v0.2.2 with default Okapi BM25 parameters. 

\paragraph{Setup.} \label{app:baseline_execution}
We compare \name with validity-oriented repair, LLM-only correction, and
generic agentic search baselines. We use three representative agentic frameworks:
\begin{itemize}
    \item \textbf{ReAct}, which interleaves reasoning with action selection from
    the current observation~\cite{yao2022react}.
    \item \textbf{ReWOO}, which constructs a structured plan before executing
    tool-based reasoning steps~\cite{xu2023rewoo}.
    \item \textbf{PlanAndAct}, which separates planning and execution during
    iterative refinement~\cite{erdogan2025plan}.
\end{itemize}

For all baseline agents, we apply SMISELF to the invalid initial molecule before
the first iteration. After each iteration, if the produced molecule is invalid,
we apply SMISELF again, enabling the agent to continue search over valid
molecular states. Each agent is additionally provided with five
BM25-retrieved reference molecules as contextual examples~\citep{robertson2009probabilistic}. ReAct and
PlanAndAct are run for five iterations, while ReWOO is run for a single
iteration following the original setting. For tool-augmented baselines, denoted
by the suffix "-T", executable molecular edit actions are used, and the agent
is allowed one retry when an action execution fails.

\paragraph{Tool Setup.}
\begin{table}[t]
\centering
\scriptsize
\setlength{\tabcolsep}{3pt}
\renewcommand{\arraystretch}{1.05}
\caption{RDKit-based molecular edit actions.}
\label{tab:rdkit_edit_actions}
\begin{tabular}{p{0.42\columnwidth}|p{0.50\columnwidth}}
\toprule
\rowcolor{gray!18}
Action & Molecular operation \\
\midrule
\texttt{replace\_substructure} &
Replace a matched SMARTS/SMILES pattern with a target fragment. \\
\texttt{delete\_substructure} &
Remove an extra matched substituent or functional group. \\
\texttt{add\_fragment} &
Attach a small fragment to the first atom matched by an attachment pattern. \\
\texttt{change\_bond\_order} &
Set the order of a matched two-atom bond. \\
\texttt{mutate\_atom} &
Change the element type of the first matched atom. \\
\texttt{set\_formal\_charge} &
Set the formal charge of the first matched atom. \\
\texttt{no\_op} &
Keep the current molecule unchanged. \\
\bottomrule
\end{tabular}
\end{table}

We provide tool-augmented baseline agents with seven executable RDKit-based
edit actions at Table \ref{tab:rdkit_edit_actions}~\citep{landrum2013rdkit}. Each action operates on the current valid molecular graph and
returns a sanitized SMILES when the edit succeeds.

\begin{itemize}
    \item \textbf{\texttt{replace\_substructure}} replaces the first matched
    source substructure with a valid target SMILES fragment. The source pattern
    may be specified as SMARTS or SMILES.

    \item \textbf{\texttt{delete\_substructure}} removes a matched substructure,
    such as an extra substituent or incorrectly attached atom, and succeeds only
    if the resulting molecule remains valid.

    \item \textbf{\texttt{add\_fragment}} attaches a valid SMILES fragment to
    the first atom matched by an attachment SMARTS pattern using a single bond.

    \item \textbf{\texttt{change\_bond\_order}} changes an existing matched bond
    to \texttt{SINGLE}, \texttt{DOUBLE}, \texttt{TRIPLE}, or
    \texttt{AROMATIC}.

    \item \textbf{\texttt{mutate\_atom}} changes the element type of the first
    atom matched by an atom SMARTS pattern, followed by sanitization.

    \item \textbf{\texttt{set\_formal\_charge}} assigns an integer formal charge
    to the first atom matched by an atom SMARTS pattern.

    \item \textbf{\texttt{no\_op}} leaves the molecule unchanged and serves as a
    safety action when further editing is unnecessary.
\end{itemize}

\section{\name : Case Reports}
We provide a qualitative case report (figure~\ref{fig:amrec_117}, \ref{fig:amrec_1865}, \ref{fig:amrec_440394}, \ref{fig:amrec_5320053}, \ref{fig:amrec_56597232}, \ref{fig:amrec_6913121}) to more closely inspect a representative example in which \name successfully recovers the intended molecule. This analysis complements the aggregate results by showing how different correction paradigms behave at the level of molecular structure.
The target description specifies a substituted flavone with multiple hydroxyl and methoxy groups. Although the initial draft is invalid, its unsanitized depiction still contains target-relevant structural cues, including the approximate flavone-like scaffold and oxygen-containing substituents. Therefore, the desired behavior is not simply to convert the draft into any valid molecule, but to recover validity while preserving and refining these chemically meaningful cues.

Validity-oriented repair methods fail to achieve this goal. SMISELF produces a valid molecule, but the repaired structure shows substantial deviation from the ground-truth molecule, yielding a very low Morgan similarity. This indicates that the method restores syntactic validity at the cost of distorting the target-relevant molecular identity. The LLM-based corrector exhibits a similar failure pattern: despite conditioning on the target description and iterative feedback, it remains close to the distorted repaired structure and does not recover the correct scaffold and substitution pattern. These results suggest that repair-based approaches can overwrite useful information already present in the invalid draft.

In contrast, \name recovers the exact target molecule in this case. \name benefits from treating the invalid draft as a corrupted molecular state rather than a broken string. Its requirement-driven \texttt{Checker}–\texttt{Critic}–\texttt{Planner} loop identifies unresolved molecule-text mismatches, while the \texttt{Candidate Explorer} maintains multiple recovery hypotheses instead of committing to a single repaired trajectory. The final selection step then chooses the candidate that best preserves the draft’s useful structural cues while satisfying the target description.
This case illustrates the central distinction between validity repair and molecular recovery. Repair methods can produce formally valid molecules, but they may introduce large structural distortions from the intended molecular identity. \name instead performs molecule-aware recovery by preserving target-relevant cues and exploring alternative trajectories before final selection.

\clearpage

\renewcommand{\thefigure}{C.\arabic{figure}}
\renewcommand{\thetable}{C.\arabic{table}}

\begin{figure}[t]
    \centering
    \includegraphics[width=\columnwidth]{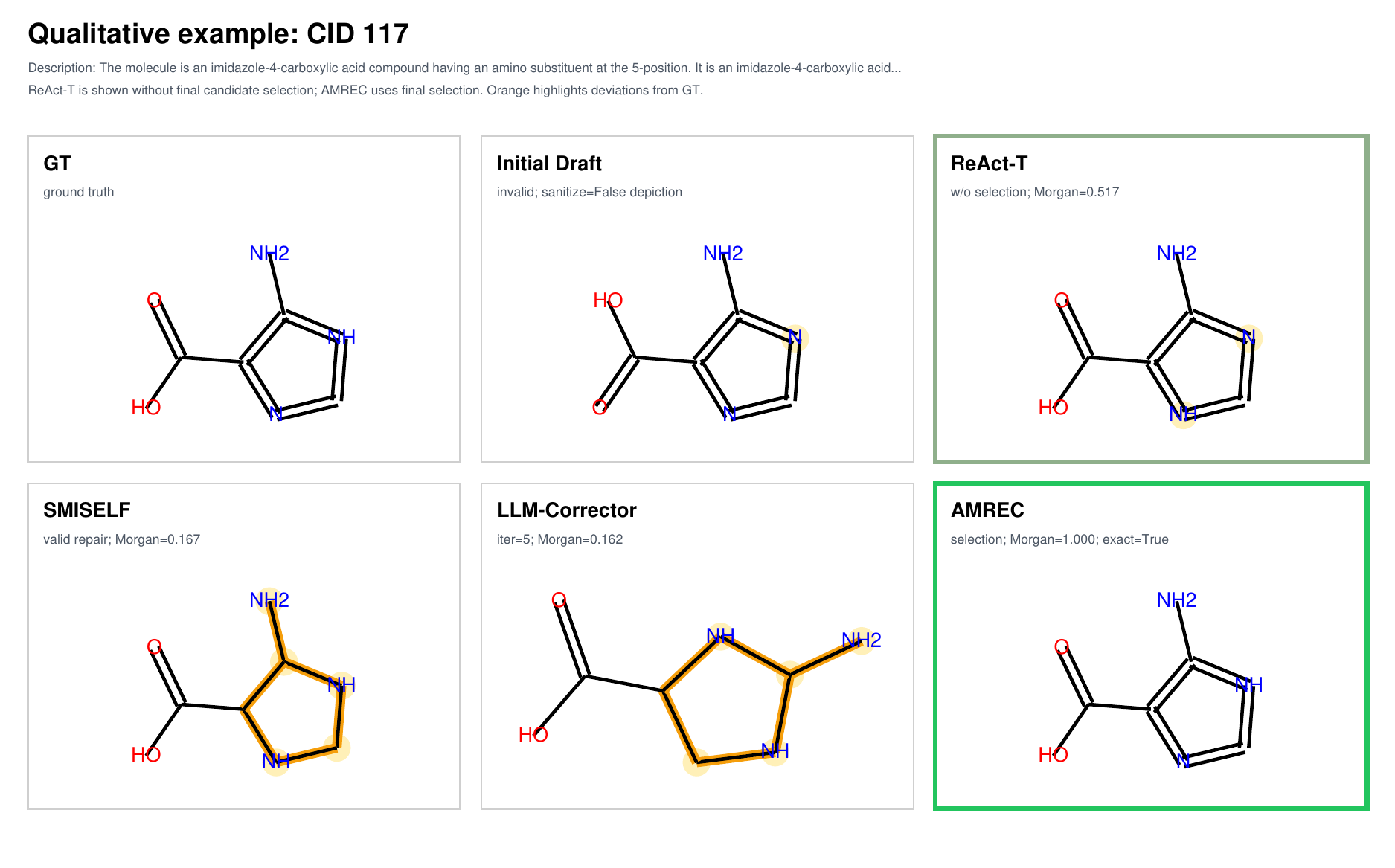} \vspace{-10mm}
    \caption{Qualitative example of molecular restoration for an invalid draft. }
    \label{fig:amrec_117} \vspace{-5mm}
\end{figure}

\begin{figure}[t]
    \centering
    \includegraphics[width=\columnwidth]{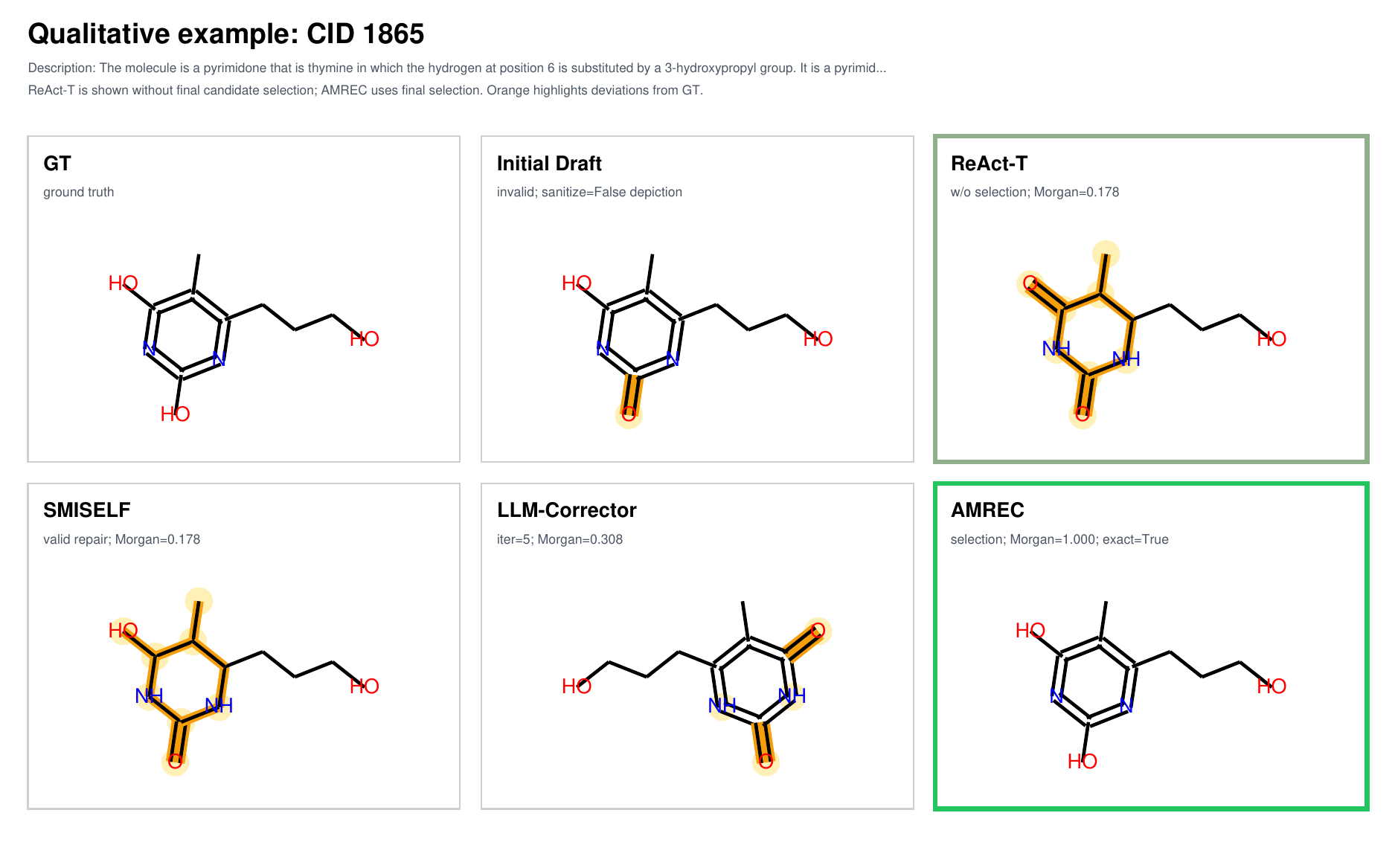} \vspace{-10mm}
    \caption{Qualitative example of molecular restoration for an invalid draft. }
    \label{fig:amrec_1865} \vspace{-5mm}
\end{figure}

\begin{figure}[t]
    \centering
    \includegraphics[width=\columnwidth]{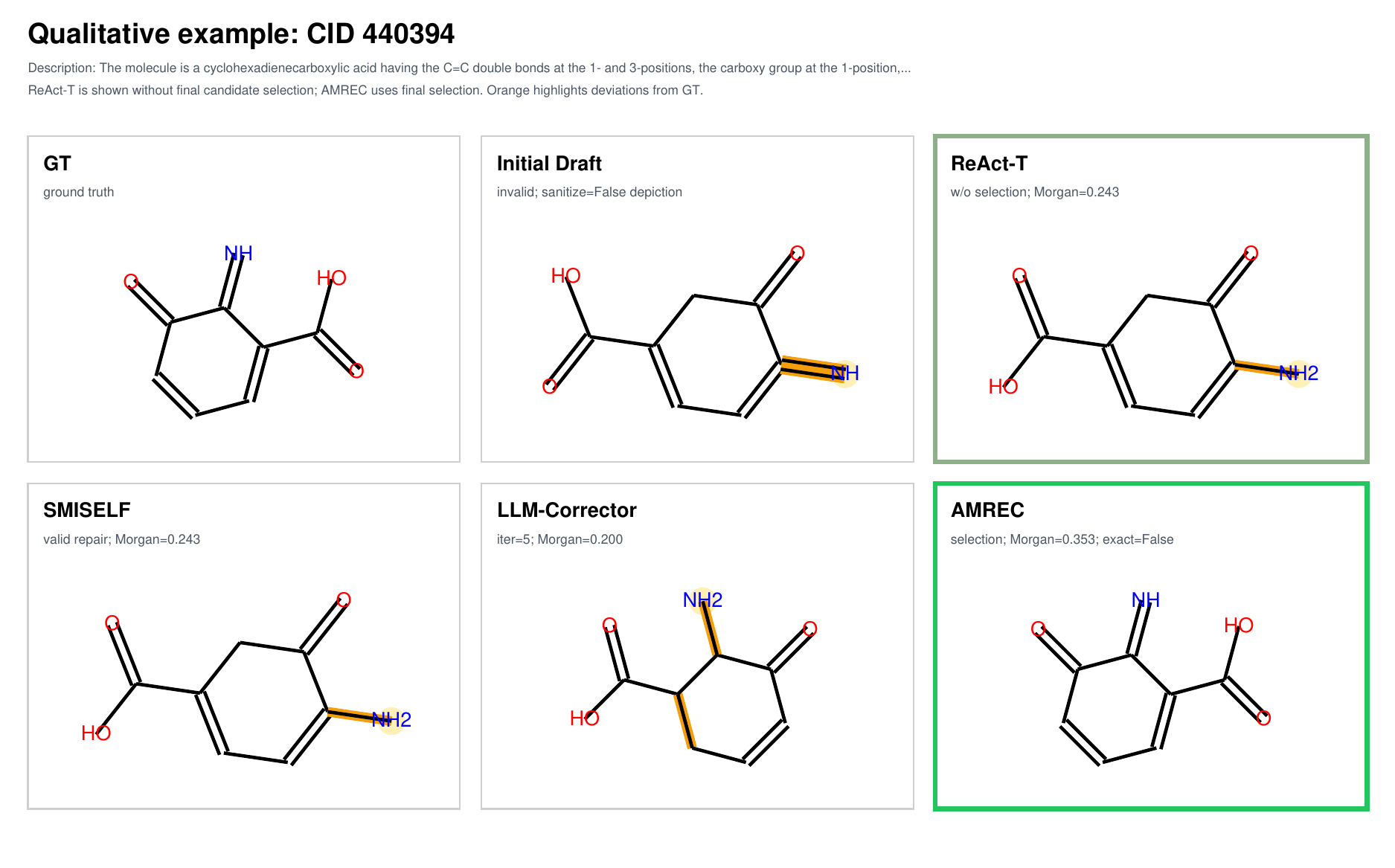} \vspace{-10mm}
    \caption{Qualitative example of molecular restoration for an invalid draft. }
    \label{fig:amrec_440394} \vspace{-5mm}
\end{figure}

\begin{figure}[t]
    \centering
    \includegraphics[width=\columnwidth]{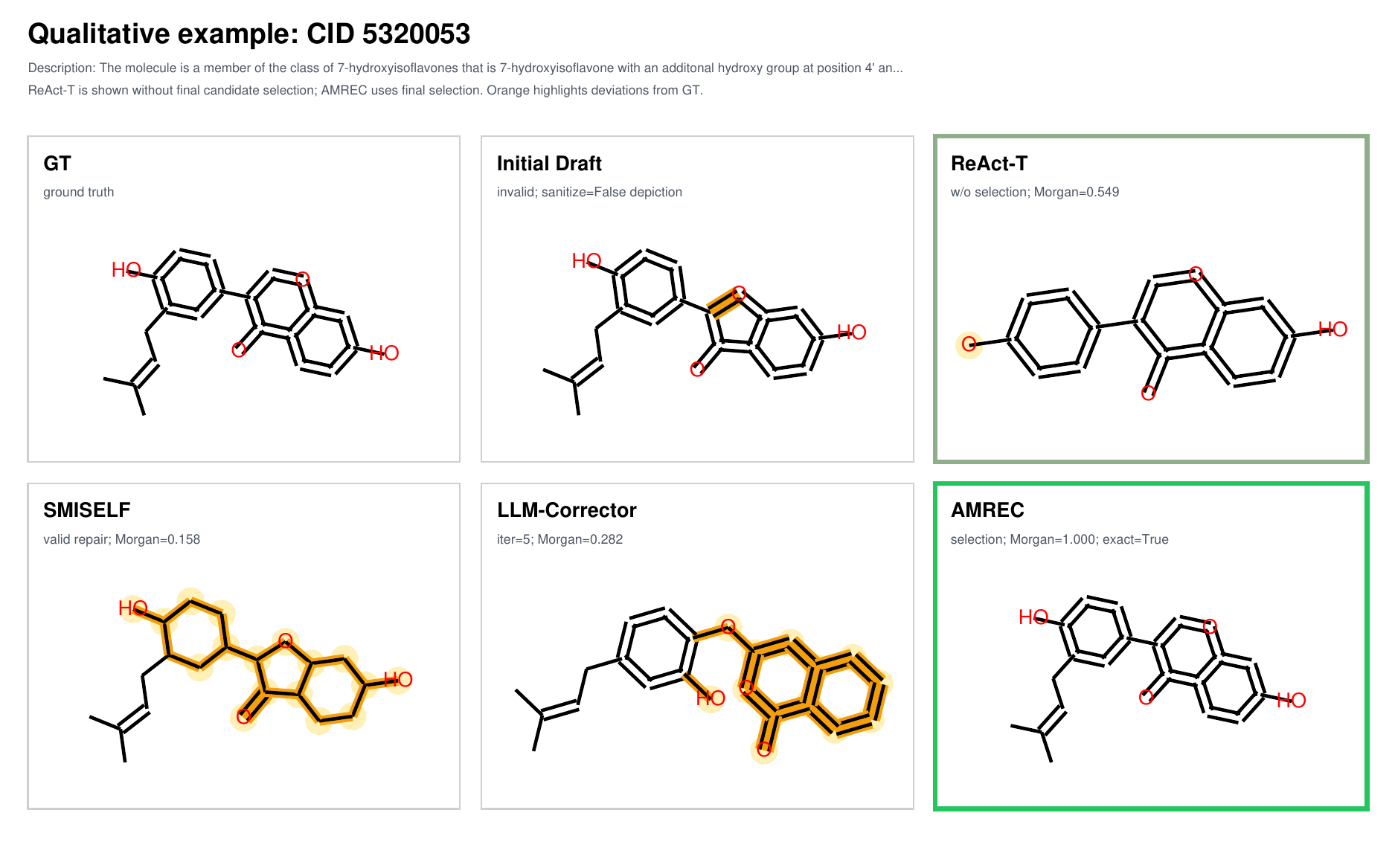} \vspace{-10mm}
    \caption{Qualitative example of molecular restoration for an invalid draft. }
    \label{fig:amrec_5320053} \vspace{-5mm}
\end{figure}

\begin{figure}[t]
    \centering
    \includegraphics[width=\columnwidth]{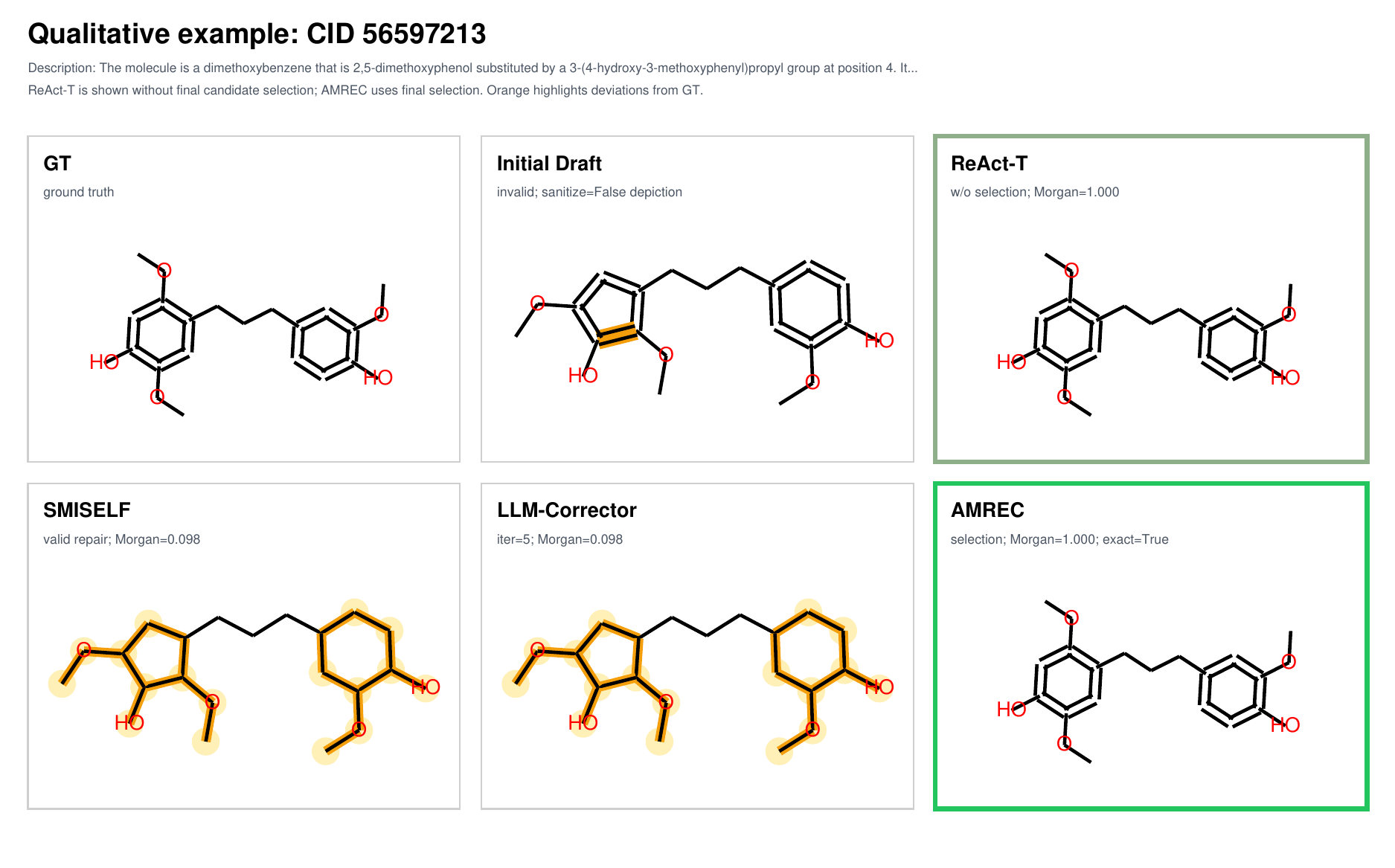} \vspace{-10mm}
    \caption{Qualitative example of molecular restoration for an invalid draft. }
    \label{fig:amrec_56597232} \vspace{-5mm}
\end{figure}

\begin{figure}[t]
    \centering
    \includegraphics[width=\columnwidth]{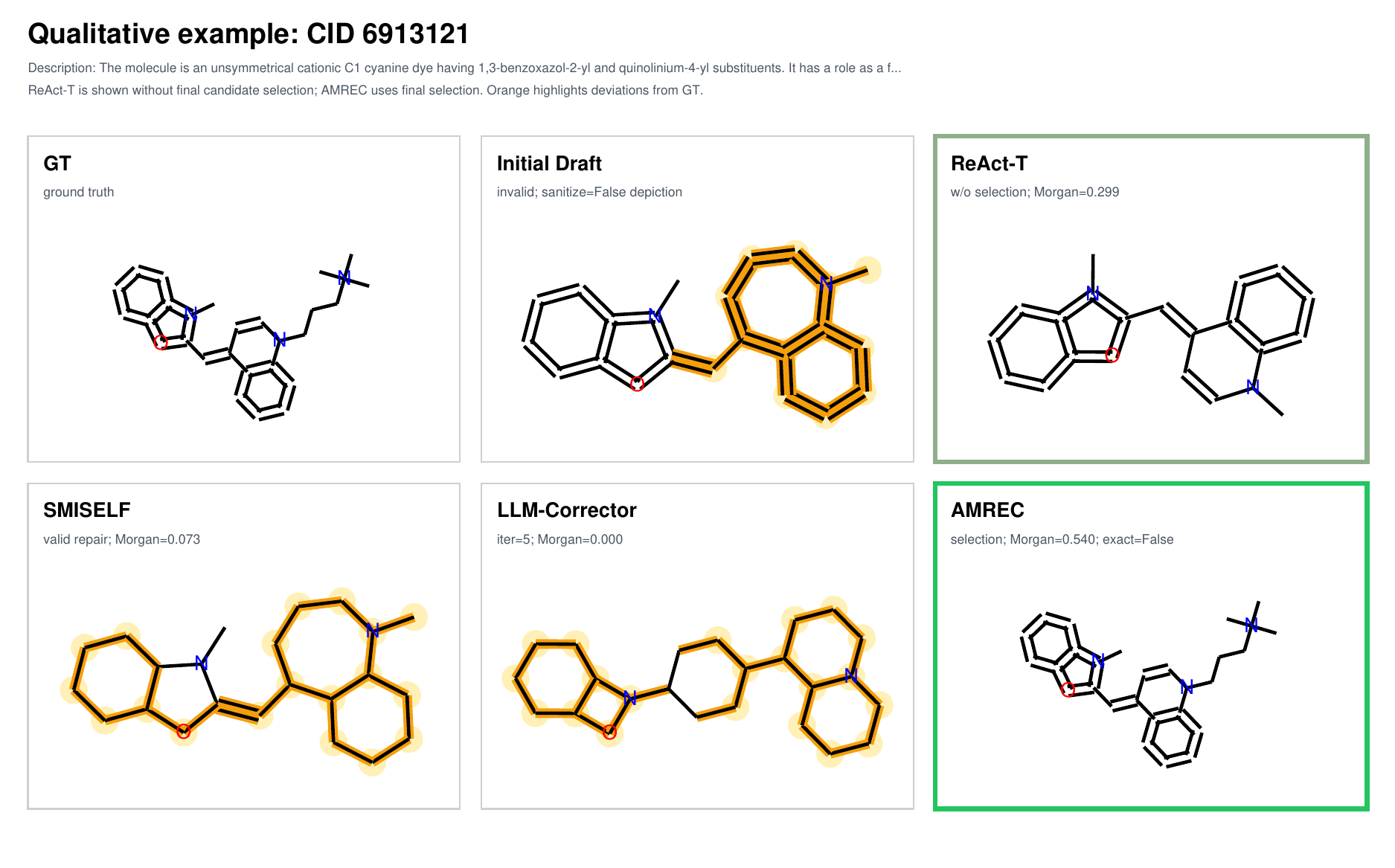} \vspace{-10mm}
    \caption{Qualitative example of molecular restoration for an invalid draft. }
    \label{fig:amrec_6913121} \vspace{-5mm}
\end{figure}
\clearpage


\newpage

\end{document}